\documentclass{article} 
\usepackage{iclr2027_conference,times}

\usepackage{amsmath,amsfonts,bm}

\def\eqref#1{equation~\ref{#1}}

\def\1{\bm{1}}

\def\ra{{\textnormal{a}}}

\def\rx{{\textnormal{x}}}

\def\rva{{\mathbf{a}}}

\def\erva{{\textnormal{a}}}

\def\ervx{{\textnormal{x}}}

\def\rmA{{\mathbf{A}}}

\def\vmu{{\bm{\mu}}}
\def\vtheta{{\bm{\theta}}}
\def\va{{\bm{a}}}

\def\ve{{\bm{e}}}

\def\vx{{\bm{x}}}

\def\eva{{a}}

\def\mA{{\bm{A}}}

\def\mH{{\bm{H}}}
\def\mI{{\bm{I}}}
\def\mJ{{\bm{J}}}

\def\mX{{\bm{X}}}

\def\mSigma{{\bm{\Sigma}}}

\DeclareMathAlphabet{\mathsfit}{\encodingdefault}{\sfdefault}{m}{sl}
\SetMathAlphabet{\mathsfit}{bold}{\encodingdefault}{\sfdefault}{bx}{n}
\newcommand{\tens}[1]{\bm{\mathsfit{#1}}}
\def\tA{{\tens{A}}}

\def\tX{{\tens{X}}}

\def\gG{{\mathcal{G}}}

\def\sA{{\mathbb{A}}}
\def\sB{{\mathbb{B}}}

\def\sS{{\mathbb{S}}}

\def\emA{{A}}

\newcommand{\etens}[1]{\mathsfit{#1}}

\def\etA{{\etens{A}}}

\newcommand{\E}{\mathbb{E}}

\newcommand{\R}{\mathbb{R}}

\newcommand{\KL}{D_{\mathrm{KL}}}
\newcommand{\Var}{\mathrm{Var}}

\newcommand{\Cov}{\mathrm{Cov}}

\newcommand{\normltwo}{L^2}
\newcommand{\normlp}{L^p}

\newcommand{\parents}{Pa} 

\usepackage{hyperref}
\usepackage{url}
\usepackage{algorithm}
\usepackage{algorithmic}
\usepackage{multirow}
\usepackage{amsmath}

\usepackage{booktabs}
\usepackage{graphicx}
\newcommand{\method}{\textsc{SCSP}}
\definecolor{cite}{HTML}{8491B4}
\usepackage{fontawesome}

\title{Selecting What Matters: Semantic Compression-Guided Selective Pooling for Long-Context Embeddings}

\author{%
  Zifeng Cheng$^\text{\textnormal{1}}$\thanks{Equal contribution.} \quad 
  Jie Zheng$^\text{\textnormal{1}}$\footnotemark[1]  \quad 
 Zhiwei Jiang$^\text{\textnormal{1}}$\thanks{Corresponding author.}  \quad 
  \textbf{Shuwen Wang}$^\text{\textnormal{1}}$ \quad
 \textbf{Fei Shen}$^\text{\textnormal{2}}$  \quad
 \textbf{Shiping Ge}$^\text{\textnormal{3}}$ \\
 \textbf{Qing Gu}$^\text{\textnormal{1}}$ \\
 $^\text{1}$ State Key Laboratory for Novel Software Technology, Nanjing University \\
$^\text{2}$ National University of Singapore
\\
$^\text{3}$ Nanjing University of Posts and Telecommunications
\\
  \texttt{chengzf@nju.edu.cn}, \texttt{231880508@smail.nju.edu.cn}, \texttt{jzw@nju.edu.cn}, \\
  \texttt{wangsw@smail.nju.edu.cn}, \texttt{shenfei29@nus.edu.sg},  \\
  \texttt{spge@njupt.edu.cn}, \texttt{guq@nju.edu.cn}
}

\iclrfinalcopy
\begin{document}

\maketitle

\lhead{Preprint}

\begin{center}
  \vspace{-30pt}
  \href{https://github.com/zifengcheng/SCSP}{\textcolor{black}{\faGithub\;}\textcolor{cite}{Github}}
\end{center}

\begin{abstract}
Large language models (LLMs) have shown strong potential as training-free text encoders for long-context embeddings.
Existing approaches primarily improve information flow under causal attention and typically construct embeddings by uniformly averaging all token representations. However, for long documents, such mean pooling can dilute salient semantic information with abundant redundant or weakly informative content.
To this end, we propose SCSP, a training-free framework that leverages semantic compression for informative token selection in long-context embedding.
Specifically, SCSP first partitions a document into sentence-aware chunks and appends a semantic compression prompt to each chunk.
A prompt-isolated attention mask preserves information flow among document tokens while restricting each prompt to its corresponding local context.
We then use the attention patterns elicited by these prompts to estimate token importance, select informative tokens, and aggregate their intermediate-layer representations into the final embedding.
Extensive experiments on long-context embedding benchmarks demonstrate that SCSP can be integrated into both zero-shot and fine-tuned models in a plug-and-play manner, consistently improving their performance.
\end{abstract}

\section{Introduction}
Text embeddings have a wide range of real-world applications, including information retrieval, text clustering, and recommender systems.
Recently, motivated by the strong long-context modeling and zero-shot generalization capabilities of large language models (LLMs), several studies~\citep{HTP,tp} have explored directly extracting long-context embeddings from LLMs without additional parameter updates.
This training-free paradigm avoids the computational cost of fine-tuning while preserving the original generative capabilities of LLMs, making off-the-shelf models readily applicable as text encoders.

Existing training-free methods improve LLM-based text embeddings by introducing backward dependencies under causal attention, allowing earlier tokens to incorporate information from subsequent positions.
Echo~\citep{echo} achieves this by repeating the input sequence and extracting representations from its second occurrence.
To avoid the computational overhead of input repetition, Token Prepending (TP)~\citep{tp} places a global semantic representation at the beginning of the input sequence.
Hierarchical Token Prepending (HTP)~\citep{HTP} further introduces multiple block-level representations, alleviating the information bottleneck caused by compressing the entire document into a single global token.

Although existing methods improve how token representations incorporate global or backward context, they largely overlook the final semantic aggregation step.
In particular, mean pooling is commonly adopted and has been shown to outperform last-token pooling for long-context embeddings~\citep{HTP}.
However, mean pooling implicitly assigns equal importance to all tokens, even though long documents often contain substantial amounts of semantically redundant or uninformative content.
Consequently, uniformly averaging all token representations may dilute the contribution of salient tokens, causing critical semantic information to be overwhelmed by less informative content~\citep{lmk}.
This observation motivates the following research question: \textit{Can we identify and selectively aggregate informative tokens, rather than treating all tokens equally, to construct long-context embeddings without additional training?}


To answer this question, we propose a training-free Semantic Compression-guided Selective Pooling (SCSP) method for long-context embedding.
Instead of treating all token representations uniformly, SCSP exploits the attention patterns elicited by semantic compression prompts as intrinsic signals of token importance.
Specifically, we partition a long document into sentence-aware chunks and append a semantic compression prompt to each chunk, enabling token importance to be estimated within focused and semantically coherent local contexts.
A prompt-isolated attention mask further prevents the inserted prompts from interfering with the original document representations while restricting each prompt to its corresponding chunk.
Based on the resulting attention distributions, SCSP selects semantically informative tokens and aggregates their representations to construct the final long-context embedding.
Extensive experiments across multiple long-context embedding benchmarks demonstrate that SCSP consistently improves existing training-free approaches and can be seamlessly integrated into fine-tuned embedding models, highlighting the general effectiveness of selective semantic aggregation.

Our main contributions are summarized as follows:
\begin{itemize}
\item We revisit the final aggregation step in long-context embedding extraction and show that selectively pooling informative tokens is more effective than uniformly averaging all tokens.

\item We propose SCSP, a training-free selective pooling framework that leverages attention elicited by local semantic compression as an intrinsic signal of token importance, enabling informative tokens to be identified and aggregated.

\item Extensive experiments across multiple long-context embedding benchmarks show that SCSP consistently improves diverse training-free baselines and further enhances fine-tuned embedding models, demonstrating the broad applicability of selective semantic aggregation.
\end{itemize}


\section{Related Work}

\paragraph{Finetuning LLMs for Embeddings.}
Text embedding models map variable-length texts into fixed-dimensional vectors and serve as fundamental components in information retrieval, clustering, and semantic text similarity.
Early neural text encoders, such as Sentence-BERT~\citep{Sentence-BERT} and SimCSE~\citep{simcse}, typically relied on metric learning to acquire semantic representations.
Recent studies~\citep{llm2vec,li2024bellm,NV-Embed} increasingly utilize LLMs as text encoders, leveraging the semantic knowledge acquired during generative pretraining. 
For example, LLM2Vec~\citep{llm2vec} converts LLMs into bidirectional encoders by removing the causal attention mask and subsequently applying masked next-token prediction and contrastive learning.
In addition, several studies \citep{jina,m2bert,mGTE,lmk} have focused on long-context embedding models.
However, these methods require large amounts of data and incur substantial fine-tuning costs, while also turning general-purpose LLMs into specialized embedding models.

\paragraph{Training-free LLMs for Embeddings.}
Motivated by the strong zero-shot capabilities of LLMs, recent studies~\citep{prompteol,cp,echo,tp,HTP} have begun to repurpose off-the-shelf LLMs as text encoders.
One line of work~\citep{prompteol,metaeol,ke,GenEOL} focuses on prompt engineering to elicit higher-quality text representations from LLMs.
A representative method, PromptEOL~\citep{prompteol}, introduces a semantic compression prompt, \textit{This sentence: ``\texttt{[TEXT]}'' means in one word: ``}, which encourages the model to compress the semantics of the input text into the representation of the final token.
Another line of work focuses on introducing backward dependencies into LLMs to overcome the limitations of causal attention.
Echo~\citep{echo} repeats the input sequence, allowing the second occurrence to attend to the entire text.
Token Prepending (TP)~\citep{tp} prepends a global semantic representation to the input sequence, while Hierarchical Token Prepending (HTP)~\citep{HTP} further introduces multiple block-level semantic representations and shows that mean pooling is more effective than last-token pooling for long-context embeddings.
Different from these approaches, our method focuses on long-context embeddings and uses the attention patterns elicited by semantic compression prompts to estimate token importance and selectively aggregate informative token representations.

\begin{figure}[t]
    \centering
    \includegraphics[width=1\textwidth]{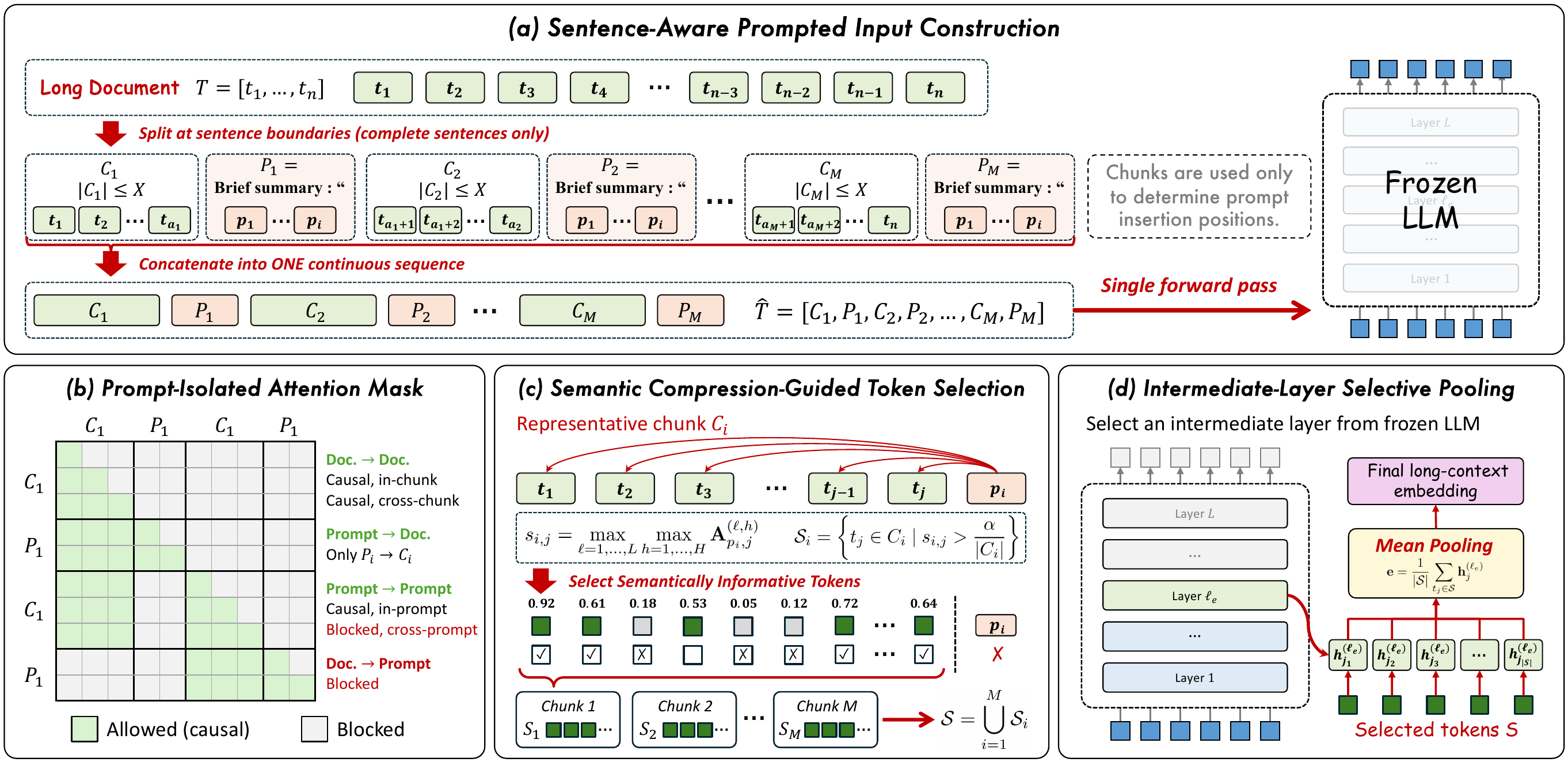}
    \caption{Illustration of the SCSP method.}
    \label{fig:framework}
\end{figure}

\section{Method}
\subsection{Overview}
The key idea of our method is to leverage chunk-wise semantic compression prompts to identify informative tokens for long-context embedding construction.
As illustrated in Figure~\ref{fig:framework}, our framework consists of four main components: input construction, prompt-isolated attention mask, semantic compression-guided token selection, and intermediate-layer selective pooling.

Given a long document, we first partition it into multiple local sentence-aware chunks and append a semantic compression prompt to each chunk.
Then, we modify the attention mask to prevent subsequent document tokens from attending to the semantic compression prompts, while restricting each prompt to attend only to tokens within its corresponding chunk.
Subsequently, we use the attention distributions elicited by the semantic compression prompts to estimate token importance and select informative tokens.
Finally, we aggregate the representations of all selected tokens from an intermediate layer to obtain the final long-context embedding.

\subsection{Input Construction}
Instead of estimating token importance over an entire long document, we first partition the input into multiple local contexts.
By reducing the semantic scope associated with each compression prompt, this partitioning enables more reliable estimation of token importance within a semantically coherent context.

Given an input text sequence $T=[t_1,\cdots,t_n]$, we partition it into $M$ non-overlapping sentence-aware chunks $T=[C_1,\cdots,C_M]$.
Each chunk consists of several complete sentences and is constrained to contain no more than $X$ tokens, where $X$ denotes the maximum chunk size.
This sentence-level partitioning preserves semantic coherence within each chunk and prevents the semantic information of a sentence from being divided across different chunks.

After partitioning the input into chunks, we augment each chunk by appending a short semantic compression prompt, $P=\texttt{Brief summary}$ : ``, to its end, inspired by PromptEOL~\citep{prompteol}.
The prompt elicits the LLM's intrinsic semantic compression capability, producing attention patterns that reflect the semantic importance of tokens within each local context.
In particular, the trailing colon and opening quotation mark in the prompt are used to prevent the model from generating punctuation as the next token, thereby enabling more effective semantic compression.

Finally, the resulting augmented input sequence is given by $ \hat{T}=[C_1, P_1, C_2, P_2, \cdots, C_M, P_M].$
Notably, chunking is used solely to determine where to insert the semantic compression prompts.
We still feed the entire augmented input sequence into the LLM at once, rather than processing each chunk separately.


\subsection{Prompt-Isolated Attention Mask}
After constructing the input, we modify the attention mask to prevent the semantic compression prompts from interfering with the original tokens in the long document and to restrict the attention scope of each prompt.

For the original document tokens, we preserve causal attention among all preceding document tokens while masking out all inserted prompt tokens.
Consequently, the document is still processed jointly as a whole, allowing information to propagate across chunk boundaries, while the inserted prompts do not directly affect the representations of the original document tokens through attention.
For the prompt appended to chunk $C_i$, we restrict its attention to the tokens in $C_i$ and the preceding tokens within the prompt itself.
Although each prompt directly attends only to its corresponding chunk, the document tokens themselves retain causal dependencies across chunk boundaries.
Therefore, each prompt serves as a local semantic readout over globally contextualized token representations.

Formally, let $\mathcal{D}$ denote the positions of the original document tokens, $\mathcal{C}_i$ denote the positions of the tokens in chunk $C_i$, and $\mathcal{P}_i$ denote the positions of the prompt appended to $C_i$. The modified attention mask matrix $\mathbf{M}$ is defined as
\[
M_{q,k} =
\begin{cases}
0, & q\in\mathcal{D},\ k\in\mathcal{D},\ k\leq q,\\
0, & q\in\mathcal{P}_i,\ k\in \mathcal{C}_i,\\
0, & q\in\mathcal{P}_i,\ k\in\mathcal{P}_i,\ k\leq q,\\
-\infty, & \text{otherwise}.
\end{cases}
\]
This design prevents information from the inserted prompts from directly propagating into the document token representations through attention, while constraining each prompt to estimate token importance within a localized semantic context.
Notably, we also explore modifying the positional encodings to keep the position indices of the original document tokens contiguous.
In this way, each document token preserves the same representation as in the unaugmented input.
However, this strategy does not yield any performance improvement.


\subsection{Semantic Compression-Guided Token Selection}

After constructing the prompt-augmented input and applying the modified attention mask, we use the attention patterns induced by the semantic compression prompts to identify informative tokens.
The intuition is that semantic compression requires the model to focus on the most informative content within each chunk, which is reflected in the attention assigned to individual tokens.
We therefore use the attention weights produced by each semantic compression prompt as estimates of token importance.

For each chunk $C_i$, let $p_i$ denote the position of the final token in its appended prompt.
Under the modified attention mask, the token at position $p_i$ can attend only to the original tokens in $C_i$ and the preceding tokens within its own prompt.
Given the normalized self-attention score matrix $\mathbf{A}^{(\ell,h)}$ of the $h$-th attention head in the $\ell$-th Transformer layer, we define the importance score of token $t_j \in C_i$ as
\[
s_{i,j}
=
\max_{\ell=1,\ldots,L}
\max_{h=1,\ldots,H}
A^{(\ell,h)}_{p_i,j},
\]
where $L$ and $H$ denote the numbers of Transformer layers and attention heads, respectively.
We take the maximum across layers and attention heads so that a strong semantic association captured by an individual layer or head can contribute to token selection.
Notably, since the normalized attention scores are used solely for token selection within each chunk, the attention assigned to the semantic compression prompt itself is excluded when normalizing the attention scores over the chunk.

We then define the selected token set $\mathcal{S}_i$ for chunk $C_i$ as
\[
\mathcal{S}_i
=
\left\{
t_j \in C_i
\mid
s_{i,j} > \frac{\alpha}{|C_i|}
\right\},
\]
where $\alpha$ is a threshold scaling factor and $|C_i|$ denotes the number of tokens in chunk $C_i$.
Since $1/|C_i|$ corresponds to uniform attention over the chunk, a token is retained if its importance score exceeds this baseline by a factor of $\alpha$.

Finally, the selected tokens from all chunks are combined into a global set for subsequent embedding construction:
\[
\mathcal{S}
=
\bigcup_{i=1}^{M} \mathcal{S}_i.
\]

\begin{table*}[t]
\small
\centering
\setlength{\tabcolsep}{5pt}
\caption{NDCG@10 (in percentage) on five datasets using Mistral-7B-Instruct-v0.3. We report the context length of 512 and an extended length of 8192. For the first four datasets that were also evaluated in HTP, $\dagger$ denotes the results reported in HTP, while $\ddagger$ denotes our reproduced results, which closely match the reported results.}\label{tab:mistral}
\resizebox{\textwidth}{!}{%
\begin{tabular}{clccccccc}
\toprule
\textbf{CXT Len} & \textbf{Method} & \textbf{QMSum} &
\textbf{2WikiMQA} & \textbf{SumFD} & \textbf{NQA} &
\textbf{MultiFieldQA} & \textbf{Avg} & \textbf{Times} \\
\midrule

\multirow{9}{*}{512}
& PromptEOL$^{\dagger}$ & 4.57 & 7.16 & 6.07 & 2.80 & 21.72 & 8.46 & -- \\
& TP w. PromptEOL$^{\dagger}$ & 5.44 & 6.51 & 5.52 & 1.72 & 17.58 & 7.35 & -- \\
& TP w. Mean$^{\dagger}$ & 11.97 & 18.62 & 38.50 & 3.08 & 44.80 & 23.39 & -- \\
\cmidrule{2-9}
& Vanilla Mean$^{\ddagger}$ & 15.01 & 16.07 & 40.39 & 2.56 & 48.61 & 24.53 & $1.00\times$ \\
& Vanilla Mean + SCSP (\textbf{\textit{Ours}}) & 14.07 & 21.21 & \textbf{45.63} & 3.27 & 52.06
& 27.25 (+2.72) & $1.19\times$ \\
\cmidrule{2-9}
& Echo Mean$^{\ddagger}$ & 14.34 & 25.63 & 32.36 & 7.06 & 60.37 & 27.95
& $1.65\times$ \\
& Echo Mean + SCSP (\textbf{\textit{Ours}}) & \textbf{16.62} & \textbf{30.06} & 40.48 &
\textbf{10.10} & \textbf{66.74} & \textbf{32.80 (+4.85)} & $1.79\times$ \\
\cmidrule{2-9}
& HTP$^{\ddagger}$ & 14.94 & 18.22 & 38.79 & 2.67 & 44.23 & 23.77
& $1.03\times$ \\
& HTP + SCSP (\textbf{\textit{Ours}}) & 14.83 & 21.79 & 44.39 & 3.14 & 46.71 & 26.17 (+2.40)
& $1.12\times$ \\

\midrule

\multirow{9}{*}{8192}
& PromptEOL$^{\dagger}$ & 6.36 & 7.49 & 9.86 & 4.08 & 26.33 & 10.82 & -- \\
& TP w. PromptEOL$^{\dagger}$ & 4.56 & 6.89 & 6.86 & 3.42 & 17.90 & 7.93 & -- \\
& TP w. Mean$^{\dagger}$ & 23.08 & 23.11 & 56.50 & 4.78 & 42.57 & 30.01 & -- \\
\cmidrule{2-9}
& Vanilla Mean$^{\ddagger}$ & 26.67 & 25.92 & 58.04 & 5.39 & 46.18 & 32.44
& $1.00\times$ \\
& Vanilla Mean + SCSP (\textbf{\textit{Ours}}) & \textbf{28.89} & 34.46 & 70.58 & 10.62 & 51.19
& \textbf{39.15 (+6.71)} & $1.12\times$ \\
\cmidrule{2-9}
& Echo Mean$^{\ddagger}$ & 17.36 & 29.49 & 39.24 & 8.75 & 53.88 & 29.74
& $2.68\times$ \\
& Echo Mean + SCSP (\textbf{\textit{Ours}}) & 26.43 & \textbf{36.65} & 55.09 & \textbf{11.22} &
\textbf{58.08} & 37.49 (+7.75) & $2.88\times$ \\
\cmidrule{2-9}
& HTP$^{\ddagger}$ & 25.36 & 25.64 & 56.38 & 5.37 & 43.88 & 31.33
& $1.05\times$ \\
& HTP + SCSP (\textbf{\textit{Ours}}) & 28.21 & 34.60 & \textbf{71.43} & 8.80 & 50.84
& 38.77 (+7.44) & $1.19\times$ \\

\bottomrule
\end{tabular}
}
\end{table*}

\subsection{Intermediate-Layer Selective Pooling}
Since final-layer representations are primarily optimized for next-token prediction, while intermediate layers tend to preserve richer semantic information~\citep{tp,DBLP:conf/acl/LiuKL024}, we use an intermediate layer as the output layer to extract token representations and aggregate them into the final long-context embedding.

Given the representation $\mathbf{h}^{(\ell_e)}_j$ of token $t_j$ from layer $\ell_e$, we obtain the final long-context embedding by mean-pooling the representations of the selected tokens:
\[
\mathbf{e}
=
\frac{1}{|\mathcal{S}|}
\sum_{t_j \in \mathcal{S}}
\mathbf{h}^{(\ell_e)}_j.
\]

\section{Experiments}

\subsection{Datasets and Experimental Settings}
We evaluate the performance of long-context embeddings on five real-world datasets from LongBench~\citep{LongBench} and LongEmbed~\citep{LongEmbed}: QMSum~\citep{QMSum}, 2WikiMultiHopQA~\citep{2Wiki}, SummScreenFD~\citep{SummScreen}, NarrativeQA~\citep{NarrativeQA}, and MultiFieldQA~\citep{LongBench}.
We use nDCG@10 as the evaluation metric.

We conduct all experiments using 4 NVIDIA Tesla V100 GPUs.
We use Spacy's parser to split the text into sentences.
We primarily evaluate \textit{LLaMA3.1-8B-Instruct}~\citep{llama} and \textit{Mistral-7B-Instruct-v0.3}~\citep{mistral} with context lengths of 512 and 8,192 tokens.
For a context length of 512, we set the chunk size to 256 and perform a grid search over the ratio values $\alpha$ in $\{1, 1.25, 1.5, 1.75, 2\}$.
For a context length of 8,192, we set the chunk size to 512 and search $\alpha$ over $\{1.5, 1.75, 2, 2.25, 2.5\}$.
For \textit{Mistral-7B-Instruct-v0.3}, following HTP~\citep{HTP}, we use the third-to-last layer as the output layer.
For \textit{LLaMA3.1-8B-Instruct}, we search for the optimal output layer among the final seven layers.

\begin{table*}[t] 
\small \centering \setlength{\tabcolsep}{5pt}
\caption{NDCG@10 (in percentage) on five datasets using LLaMA-3.1-8B-Instruct.
}\label{tab:llama}
\resizebox{0.98\linewidth}{!}{
\begin{tabular}{clccccccc}
\toprule
\textbf{CXT Len} & \textbf{Method} & \textbf{QMSum} &
\textbf{2WikiMQA} & \textbf{SumFD} & \textbf{NQA} &
\textbf{MultiFieldQA} & \textbf{Avg} & \textbf{Times} \\
\midrule

\multirow{9}{*}{512}
& PromptEOL$^{\dagger}$ & 10.17 & 15.20 & 11.39 & 5.22 & 46.70 & 17.73 & -- \\
& TP w. PromptEOL$^{\dagger}$ & 8.33 & 8.29 & 10.97 & 1.91 & 25.65 & 11.03 & -- \\
& TP w. Mean$^{\dagger}$ & 7.13 & 5.29 & 6.95 & 1.57 & 17.72 & 7.73 & -- \\
\cmidrule{2-9}
& Vanilla Mean$^{\ddagger}$ & 12.21 & 18.09 & 35.92 & 2.82 & 45.39 & 22.89
& $1.00\times$ \\
& Vanilla Mean + SCSP (\textbf{\textit{Ours}}) & 11.60 & 26.08 & 48.57 & 3.48 & 49.86
& 27.92 (+5.03) & $1.34\times$ \\
\cmidrule{2-9}
& Echo Mean$^{\ddagger}$ & 13.86 & 22.85 & 36.35 & 7.56 & 67.83 & 29.69
& $1.90\times$ \\
& Echo Mean + SCSP (\textbf{\textit{Ours}}) & \textbf{13.68} & \textbf{28.02} &
\textbf{50.79} & \textbf{9.15} & \textbf{72.93} &
\textbf{34.91 (+5.22)} & $2.27\times$ \\
\cmidrule{2-9}
& HTP$^{\ddagger}$ & 10.37 & 8.64 & 27.81 & 2.22 & 27.21 & 15.25
& $1.01\times$ \\
& HTP + SCSP (\textbf{\textit{Ours}}) & 11.54 & 13.58 & 30.91 & 2.28 & 30.53
& 17.77 (+2.52) & $1.38\times$ \\

\midrule

\multirow{9}{*}{8192}
& PromptEOL$^{\dagger}$ & 8.16 & 15.86 & 20.47 & 7.40 & 43.85 & 19.15 & -- \\
& TP w. PromptEOL$^{\dagger}$ & 4.95 & 9.87 & 24.19 & 2.42 & 19.81 & 12.25 & -- \\
& TP w. Mean$^{\dagger}$ & 10.00 & 8.75 & 17.95 & 2.66 & 19.28 & 11.73 & -- \\
\cmidrule{2-9}
& Vanilla Mean$^{\ddagger}$ & 27.04 & 19.76 & 54.39 & 7.50 & 47.83 & 31.30
& $1.00\times$ \\
& Vanilla Mean + SCSP (\textbf{\textit{Ours}}) & \textbf{29.24} & 33.27 & \textbf{76.66} &
12.15 & 51.62 & 40.59 (+9.28) & $1.08\times$ \\
\cmidrule{2-9}
& Echo Mean$^{\ddagger}$ & 18.00 & 17.62 & 38.70 & 8.14 & 49.57 & 26.41
& $2.74\times$ \\
& Echo Mean + SCSP (\textbf{\textit{Ours}}) & 23.52 & \textbf{39.28} & 62.38 &
\textbf{17.27} & \textbf{69.26} & \textbf{42.34 (+15.94)}
& $2.92\times$ \\
\cmidrule{2-9}
& HTP$^{\ddagger}$ & 15.73 & 8.93 & 41.48 & 2.97 & 26.97 & 19.21
& $1.05\times$ \\
& HTP + SCSP (\textbf{\textit{Ours}}) & 21.49 & 14.66 & 44.88 & 4.58 & 29.61
& 23.05 (+3.83) & $1.22\times$ \\
\bottomrule
\end{tabular}
}
\end{table*}

\subsection{Baselines}
We compare our method with the following baselines.
\textbf{PromptEOL}~\citep{prompteol} compresses the semantics of a text into the last token to extract embeddings.
\textbf{TP}~\citep{tp} prepends a global semantic token to the beginning of the input.
For TP, we also explore two embedding extraction strategies: \textbf{TP w. PromptEOL}, which uses last-token pooling, and \textbf{TP w. Mean}, which uses mean pooling.
\textbf{Vanilla Mean} directly applies mean pooling over the hidden states of all tokens produced by the LLM.
\textbf{Echo Mean}~\citep{echo} repeats the text twice and uses mean-pooling to extract embeddings from the second occurrence.
\textbf{HTP}~\citep{HTP} extends TP by introducing multiple block-level representations, and constructs the final embedding through mean pooling over all token representations.
In particular, to ensure a fair comparison, all baseline methods use the same output layer for representation extraction.

\subsection{Results}
As shown in Tables~\ref{tab:mistral} and~\ref{tab:llama}, our method consistently improves the performance of Vanilla Mean, Echo Mean, and HTP across both Mistral-7B-Instruct-v0.3 and LLaMA-3.1-8B-Instruct models.
These results demonstrate that our method is effective and can be seamlessly integrated with baselines in a plug-and-play manner.
Using LLaMA with a context length of 8,192, combining our method with Echo Mean improves the average performance by 60\% relative to Echo Mean alone.
Notably, when combined with our method, the simplest Vanilla Mean achieves the best performance on Mistral-7B-Instruct-v0.3 at a context length of 8,192.

At a context length of 512, Echo Mean serves as a strong baseline because input repetition enhances the model's contextual understanding. Our method is complementary to Echo Mean and further improves its performance.
At a context length of 8,192, Echo Mean generally fails to yield performance improvements, possibly because the context becomes excessively long.
In contrast, our method achieves larger gains in this setting, likely because selectively preserving salient semantic information becomes increasingly important as the context length grows.

Notably, compared with the substantial runtime overhead incurred by Echo's input repetition, our method remains lightweight and introduces only a marginal additional runtime cost.

\begin{table*}[t] 
\small \centering \setlength{\tabcolsep}{8pt}
\caption{Ablation study results on Mistral-7B-Instruct-v0.3 with an 8,192-token context length.}\label{tab:abaltion}
    \resizebox{0.98\linewidth}{!}{
        \begin{tabular}{lcccccc}
        \toprule
         \textbf{Method} & \textbf{QMSum} & \textbf{2WikiMQA} & \textbf{SumFD} & \textbf{NQA}  & \textbf{MultiFieldQA} & \textbf{Avg} \\
        \midrule
         Vanilla Mean &  26.67 & 25.92 & 58.04 & 5.39 & 46.18 & 32.44\\
        Vanilla Mean + SCSP (\textbf{\textit{Ours}}) & 28.89 & 34.46 &\textbf{70.58} & 10.62 & 51.19 & \textbf{39.15}  \\
        \midrule
        \textit{Input Construction:}\\
        w/o Sentence-aware Chunking  & 28.65 & 33.47 & 69.17 & 8.27 & 51.13 & 38.14 \\
        w/o Chunk-wise Prompt & 24.59 & 29.56 & 67.25 & 8.43 &\textbf{52.30}  &36.42 \\
        w/o Semantic Compression Prompt & 26.17 & 29.08 & 59.50 & 8.59 & 48.55 & 34.38\\
        \midrule
        \textit{Prompt Isolation and Position Encoding:}\\
        w/o Prompt-Isolated Attention Mask &27.90&29.61	&61.41	&8.40 &50.96 &35.66 \\
        w/ Positional Encoding Modification & \textbf{29.05} & 34.60 & 69.14 & 10.07 & 51.13 & 38.80 \\
        \midrule
        \textit{Token Selection and Aggregation:}\\
        Random Token Selection  & 26.22 & 27.88 & 57.51 & 5.79 & 45.91 & 32.66\\
        Aggregation of Prompt-Compressed Representations & 26.03 & 26.93 & 59.70 & 10.43 & 47.56 & 34.13\\
        Importance-Weighted Token Aggregation & 28.85 & \textbf{34.80} & 68.86 & \textbf{10.80} & 50.83 & 38.83\\
        \bottomrule
        \end{tabular}
}
\end{table*}

\subsection{Ablation Study}
We conduct ablation studies to investigate the effects of key design choices in SCSP, as shown in Table~\ref{tab:abaltion}.

We investigate the effect of input construction on performance.
First, we replace the sentence-aware semantic chunking strategy with fixed-length chunking.
This variant leads to an average performance drop of 1.01\%, demonstrating that sentence-level chunking better preserves semantic coherence and facilitates effective token selection.
Second, instead of appending a semantic compression prompt to each chunk, we append a single prompt to the end of the entire long text for token selection.
This modification results in a performance drop of 2.73\%, highlighting the importance of performing token selection within localized chunk-level contexts.
This may be because the input text is excessively long and attention scores are influenced by the relative distances between tokens.
Finally, we ablate the prompt and directly use the attention scores from the final token of each chunk to the preceding tokens for token selection.
The performance decreases by 4.77\%, indicating that using semantic compression prompts to select tokens is a more effective strategy.

We investigate the effect of prompt isolation and position encoding on performance.
First, we retain the original causal attention pattern without modifying the attention mask, which results in a 3.49\% performance decrease.
This finding indicates that allowing tokens to attend to the additionally appended prompts interferes with their representations and consequently degrades the quality of long-context embeddings.
Second, we explore modifying the positional encodings to keep the position indices of all document tokens contiguous, which results in a 0.35\% performance drop.
This may be because retaining the positional gaps introduced by the inserted prompts helps the model better distinguish and capture semantic information within each chunk.

We further investigate token selection strategies and aggregation methods. First, we randomly select the same number of tokens within each chunk. The performance is close to that of Vanilla Mean, indicating that the gains of our method do not simply come from aggregating fewer tokens, but from identifying tokens that capture the core semantics of long documents.
Second, we directly aggregate the representation of the final token in the semantic compression prompt for each chunk, which serves as the compressed semantic representation of that chunk. This modification leads to a 5.02\% performance drop.
Consistent with the findings of HTP~\citep{HTP}, this result suggests that aggregating token-level representations is more effective than directly aggregating compressed prompt representations.
Finally, we use the token importance scores as weights to perform weighted aggregation of token representations.
We observe no further improvement over SCSP, suggesting that token importance scores are effective for token selection but are less suitable as weights for representation aggregation.

\begin{figure*}[t]
    \centering
    \includegraphics[width=1\textwidth]{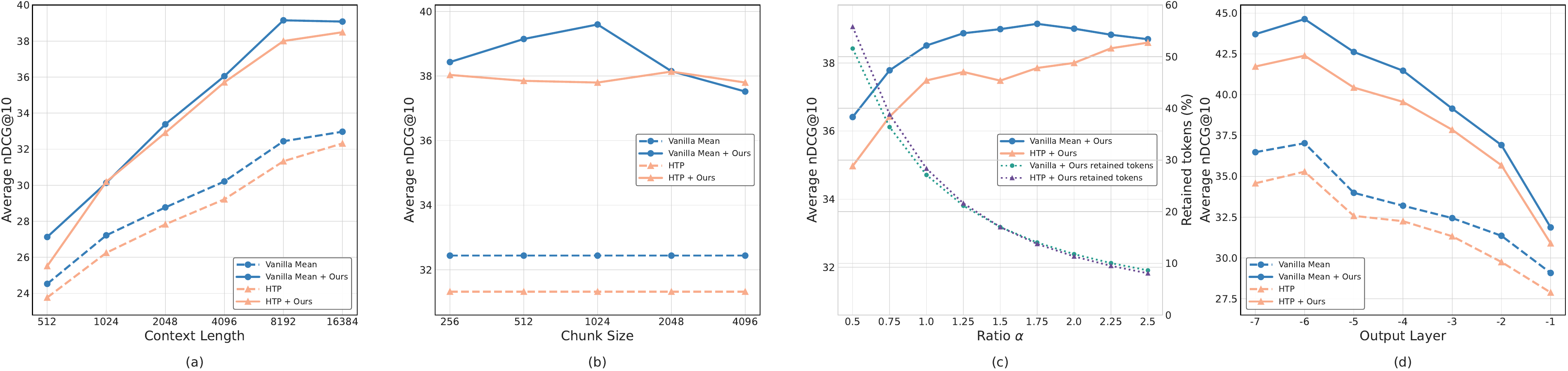}
    \caption{The average effects of context length, chunk size, ratio $\alpha$, and output layer on performance across five datasets are evaluated using Mistral-7B-Instruct-v0.3.}
    \label{fig:analysis}
\end{figure*}

\begin{table*}[t] \small \centering \setlength{\tabcolsep}{5pt}
\caption{Generalization across different LLM backbones on five datasets with an 8,192-token context length.} \label{tab:backbone}
        \resizebox{0.98\linewidth}{!}{
        \begin{tabular}{clcccccc}
        \toprule
        \textbf{Backbone} & \textbf{Method} & \textbf{QMSum} & \textbf{2WikiMQA} & \textbf{SumFD} & \textbf{NQA}  & \textbf{MultiFieldQA} & \textbf{Avg} \\
        \midrule
        \multirow{2}{*}{Gemma2-9B} 
        & Vanilla Mean &  30.78	&33.82	&70.55	&7.95	&49.99	&38.62\\
        & Vanilla Mean + SCSP (\textbf{\textit{Ours}}) & \textbf{32.72} &\textbf{43.56} &\textbf{73.61}	&\textbf{11.55}	&\textbf{53.35}	&\textbf{42.96 (+4.34)} \\
        \midrule
        \multirow{2}{*}{Qwen2.5-7B-Instruct}
        & Vanilla Mean &  25.55	&22.43	&47.49	&6.33	&\textbf{50.16}	&30.39\\
        & Vanilla Mean + SCSP (\textbf{\textit{Ours}}) & \textbf{26.66}	&\textbf{25.90} &\textbf{54.77} &\textbf{7.90} &45.52 &\textbf{32.15 (+1.76)} \\
        \bottomrule
        \end{tabular}
}
\end{table*}

\subsection{Generalizability across Different Context Lengths}
We further investigate the effect of different context lengths in Figure~\ref{fig:analysis}(a).
First, our method consistently improves the performance of both baseline methods across multiple context lengths, demonstrating its effectiveness. Second, larger context lengths generally lead to better performance. However, the performance gains gradually diminish as the context window increases.
Notably, at a context length of 16,384, Vanilla Mean achieves further performance improvements on SumFD and NQA, suggesting that these two datasets benefit from a larger context window.

\subsection{Effects of Hyperparameters}
We further investigate the effects of different hyperparameters in Figure~\ref{fig:analysis}(b), ~\ref{fig:analysis}(c), and \ref{fig:analysis}(d).
Our method consistently improves the performance of both Vanilla Mean and HTP across a wide range of hyperparameter settings, demonstrating its effectiveness and robustness.

For \textit{chunk size}, Vanilla Mean achieves its best performance with a chunk size of 1,024, whereas HTP performs best with a chunk size of 2,048.
This suggests that the optimal chunk size varies across baselines, although our method yields consistent improvements for both.
For the \textit{ratio}, Vanilla Mean achieves the best performance at 1.75, while HTP performs best at 2.5.
One possible explanation is that the token representations used by HTP incorporate more global semantic information and are therefore more contextualized, allowing a higher selection threshold to be applied.
Notably, under their respective best-performing configurations, SCSP retains only 14.03\% of the tokens for Vanilla Mean and 8.64\% for HTP. This result indicates that aggregating representations from a small subset of informative tokens is sufficient to construct effective long-context embeddings.
For the \textit{output layer}, the best performance is obtained at the sixth-to-last layer, after which the performance gradually decreases as the output layer moves closer to the final layer.

\subsection{Generalization across Different Backbones}
We further report the performance of vanilla mean on Gemma2-9B~\citep{gemma} and Qwen-2.5-7B-Instruct~\citep{qwen2.5} to validate the generalizability of our method across different backbones.

Experimental results in Table~\ref{tab:backbone} show that our method is also effective on these two LLMs, further demonstrating its generalizability.
Among the four LLMs, Gemma2-9B achieves the best performance with Vanilla Mean at a context length of 8,192.

\begin{table*}[t] 
\scriptsize
\centering
\setlength{\tabcolsep}{9pt}
\caption{Comparison of different strategies for computing token importance scores on Mistral-7B-Instruct-v0.3 with an 8,192-token context length.}
\label{tab:score}
\begin{tabular}{lcccccccc}
\toprule
\textbf{Method} & \textbf{Layer} & \textbf{Head} &
\textbf{QMSum} & \textbf{2WikiMQA} & \textbf{SumFD} &
\textbf{NQA} & \textbf{MultiFieldQA} & \textbf{Avg} \\
\midrule
Vanilla Mean & -- & -- &
26.67 & 25.92 & 58.04 & 5.39 & 46.18 & 32.44 \\
\midrule
\multirow{4}{*}{SCSP}
& Max  & Max  & 28.89 & \textbf{34.46} & \textbf{70.58} & 10.62 & \textbf{51.19} & \textbf{39.15} \\
& Max  & Mean &   \textbf{29.35}    &  34.53     &  69.06     &   9.91    &  50.87     &  38.75     \\
& Mean & Max  &   27.58    &  33.20     &  67.38     &  \textbf{10.69}     &  50.57     &  37.89     \\
& Mean & Mean &   25.91  &  30.30   &  63.08  &  10.10  &  49.15     &  35.71     \\
\bottomrule
\end{tabular}
\end{table*}

\begin{table*}[t] \small \centering
\setlength{\tabcolsep}{6pt}
\caption{Generalization of SCSP to fine-tuned embedding models on five datasets with an 8,192-token context length.}  \label{tab:finetuned}
\resizebox{0.98\linewidth}{!}{
\begin{tabular}{llcccccc}
\toprule
\textbf{Models} & \textbf{Method} & \textbf{QMSum} & \textbf{2WikiMQA} & \textbf{SumFD} & \textbf{NQA} & \textbf{MultiFieldQA} & \textbf{Avg} \\
\midrule
\textbf{GritLM} & Vanilla Mean & 19.98 & 27.63 & 29.48 & 5.89 & 72.46 & 31.09 \\
              & Vanilla Mean + SCSP (\textbf{\textit{Ours}}) & \textbf{24.34} & \textbf{37.27} & \textbf{57.41} & \textbf{14.74} & \textbf{78.31} & \textbf{42.42 (+11.33)} \\
\midrule
\textbf{NV-EMBED} & Vanilla Mean & 24.87 & 34.77 & 67.11 & 24.77 & 74.61 & 45.23 \\
                & Vanilla Mean + SCSP (\textbf{\textit{Ours}}) & \textbf{28.12} & \textbf{34.99} & \textbf{72.55} & \textbf{29.15} & \textbf{76.63} & \textbf{48.29 (+3.06)} \\
\bottomrule
\end{tabular}
}
\end{table*}

\subsection{Effects of Token Importance Scores}
We further investigate the effect of different strategies for computing token importance scores in Table~\ref{tab:score}. All four strategies improve performance, with taking the maximum across all layers and attention heads achieving the best results, while averaging across all layers and heads performs the worst.
This suggests that the max operation is more effective at uncovering salient semantic signals captured by individual layers or attention heads, whereas mean aggregation may smooth out such informative signals.

\subsection{Results on Finetuned Embedding Models}
We further investigate whether our method is also effective for fine-tuned embedding models in Table~\ref{tab:finetuned}.
Specifically, we replace their mean pooling operation with our selective pooling method and use the final-layer representations to construct the embeddings.
The results show that our method consistently improves the performance of both GritLM~\citep{gritlm} and NV-EMBED~\citep{NV-Embed}.
Notably, NV-EMBED employs latent attention to update each token representation.
This suggests that semantically irrelevant tokens remain unable to capture useful semantic information even after latent-attention refinement.
These results demonstrate that our method is effective not only for general-purpose LLMs but also for further improving fine-tuned embedding models, highlighting its generalizability.

\section{Conclusion}
In this work, we introduced SCSP, a training-free framework for constructing long-context embeddings through semantic compression-guided selective pooling.
SCSP first partitions a long document into sentence-aware chunks and appends a semantic compression prompt to each chunk to estimate token importance.
A prompt-isolated attention mask prevents the inserted prompts from interfering with the original document representations, while the resulting attention patterns are used to identify and selectively aggregate semantically informative tokens.
Extensive experiments demonstrate that SCSP consistently improves Vanilla Mean, Echo Mean, and HTP across different context lengths and LLM backbones.
Moreover, SCSP also enhances finetuned embedding models, demonstrating its broad applicability.
These results highlight selective token aggregation as an effective alternative to conventional mean pooling.

\subsection*{AI use statement}

Generative AI tools were used to assist language editing, translation, and LaTeX preparation.
The authors take responsibility for the final content, including all AI-assisted text and artifacts.

\bibliography{iclr2027_conference}
\bibliographystyle{iclr2027_conference}

\clearpage
\appendix
\section{Effects of Semantic Compression Prompt}
We investigate the effects of different prompts on performance in Table~\ref{tab:prompt}.

All evaluated prompts yield performance improvements, indicating that our method generalizes well across different prompts and exhibits broad applicability.
We further remove the opening quotation mark from the prompt and observe a performance decrease of 0.77\%.
This finding suggests that the opening quotation mark provides a more effective cue for eliciting the desired semantic continuation.

\begin{table*}[t]
\caption{
Effects of semantic compression prompts on Mistral-7B-Instruct-v0.3 with an 8,192-token context length.
}
\small
\centering
\setlength{\tabcolsep}{12pt}
\resizebox{0.98\linewidth}{!}{
\begin{tabular}{lcccccc}
\toprule
\textbf{Prompt}
& \textbf{QMSum}
& \textbf{2WikiMQA}
& \textbf{SumFD}
& \textbf{NQA}
& \textbf{MultiFieldQA}
& \textbf{Avg} \\
\midrule
Brief summary: ``
& \textbf{28.89} & 34.46 & \textbf{70.58}
& \textbf{10.62} & 51.19 & \textbf{39.15} \\

Brief summary:
& 28.62 & 34.14 & 67.59
& 10.03 & 51.54 & 38.38 \\

This text means in one word: ``
& 28.22 & 31.88 & 65.24 & 9.41 & 50.98 & 37.15 \\

This text can be summarized as: ``
& 28.01 & 33.72 & 65.92 & 9.5 & \textbf{52.23} & 37.87 \\

Summary: ``
& 28.42 & \textbf{35.05} & 70.07 & 10.05 & 52.15 & \textbf{39.15} \\
\bottomrule
\end{tabular}
\label{tab:prompt}
}
\end{table*}

\section{Results on the LOCOV1 Dataset}

We further evaluate our method on the LOCOV1 dataset~\citep{m2bert}. 

SCSP consistently improves Vanilla Mean across all eight datasets, increasing the average score from 55.77 to 60.77, with an absolute gain of 5.00 points.
The improvements are particularly pronounced on SumFD and QASPER Title, where SCSP yields gains of 12.56 and 9.13 points, respectively.
Consistent gains are also observed across QA, retrieval, and long-document understanding tasks, indicating that selective aggregation can effectively suppress less informative tokens and produce more discriminative long-context representations.

\begin{table}[t]
\small
\centering
\setlength{\tabcolsep}{8pt}
\caption{Results on the LOCOV1 dataset using Mistral-7B-Instruct-v0.3 with an 8,192-token context length.}
\label{tab:locov1}
\begin{tabular}{lcc}
\toprule
\textbf{Dataset} & \textbf{Vanilla Mean} & \textbf{Vanilla Mean + SCSP (\textit{Ours})} \\
\midrule
SumFD       & 57.68 & \textbf{70.24} \\
Gov. Report & 97.30 & \textbf{98.12} \\
QMSUM       & 50.94 & \textbf{55.12} \\
QASPER  Title    & 37.27 & \textbf{46.40} \\
QASPER  Abstract    & 89.19 & \textbf{93.45} \\
2WikiMQA    & 46.90 & \textbf{51.76} \\
Passage Retrieval & 49.05 & \textbf{52.99}\\
CourtListener - Plain Text &  17.85 & \textbf{18.04} \\
\midrule
Avg.        &  55.77 & \textbf{60.77 (+5.00)} \\
\bottomrule
\end{tabular}
\end{table}

\end{document}